\pdfoutput=1

\documentclass[conference]{IEEEtran}
\IEEEoverridecommandlockouts
\usepackage{cite}
\usepackage{amsmath,amssymb,amsfonts}
\usepackage{algorithmic}
\usepackage{graphicx}

\usepackage{caption}
\usepackage{textcomp}
\usepackage{xcolor}
\usepackage{multirow}
\usepackage{booktabs}
\usepackage[linesnumbered, ruled]{algorithm2e}
\usepackage{bm}
\usepackage{color}
\usepackage[section]{placeins}
\SetKwRepeat{Do}{do}{while}%

\newcommand{\x}{{\bf x}}

\newcommand{\Ocal}{\mathcal{O}}

\renewcommand{\xi}{\x}
\renewcommand{\eta}{f}

\def\BibTeX{{\rm B\kern-.05em{\sc i\kern-.025em b}\kern-.08em
    T\kern-.1667em\lower.7ex\hbox{E}\kern-.125emX}}

\begin{document}

\makeatletter
\newcommand{\linebreakand}{%
\end{@IEEEauthorhalign}
\hfill\mbox{}\par
\mbox{}\hfill\begin{@IEEEauthorhalign}
}
\makeatother

%\title{Autoencoder and Linear Model of Coregionalization for the Prediction of High-Dimensional Outputs \\
% \thanks{* Xueying Zhang and Wei Xing are the corresponding authors.}
%}

\title{E-LMC: Extended Linear Model of Coregionalization for Spatial Field Prediction}

\author{\IEEEauthorblockN{Anonymous Authors}}
\author{
\IEEEauthorblockN{Shihong Wang}
\IEEEauthorblockA{\textit{School of Integrated Circuit Science and Engineering} \\
\textit{Beihang University}\\
Beijing, China \\
wangshihong@buaa.edu.cn}
\and
\IEEEauthorblockN{Xueying Zhang}
\IEEEauthorblockA{\textit{School of Integrated Circuit Science and Engineering} \\
\textit{Beihang University}\\
Beijing, China \\
xueying.zhang@buaa.edu.cn}
\linebreakand
\IEEEauthorblockN{Yichen Meng}
\IEEEauthorblockA{\textit{School of Integrated Circuit Science and Engineering} \\
\textit{Beihang University}\\
Beijing, China \\
mengyc@buaa.edu.cn}
\and
\IEEEauthorblockN{Wei W. Xing* \thanks{*Corresponding author.}}
\IEEEauthorblockA{\textit{School of Integrated Circuit Science and Engineering} \\
\textit{Beihang University}\\
Beijing, China \\
wxing@buaa.edu.cn}
}

\maketitle

\begin{abstract}
Physical simulations based on partial differential equations typically generate spatial fields results, which are utilized to calculate specific properties of a system for engineering design and optimization.
Due to the intensive computational burden of the simulations, a surrogate model mapping the low-dimensional inputs to the spatial fields are commonly built based on a relatively small dataset. 
To resolve the challenge of predicting the whole spatial field, the popular linear model of coregionalization (LMC) can disentangle complicated correlations within the high-dimensional spatial field outputs and deliver accurate predictions.
However, LMC fails if the spatial field cannot be well approximated by a linear combination of base functions with latent processes.
In this paper, we present the Extended Linear Model of Coregionalization (E-LMC) by introducing an invertible neural network to linearize the highly complex and nonlinear spatial fields so that the LMC can easily generalize to nonlinear problems while preserving the traceability and scalability.
Several real-world applications demonstrate that E-LMC can exploit spatial correlations effectively, showing a maximum improvement of about 40\% over the original LMC and outperforming the other state-of-the-art spatial field models.
\end{abstract}


\section{Introduction}
Applications such as design optimization and inverse parameter estimation require repeated solutions to sophisticated partial differential equations (PDEs) \cite{bilionis2013multi, keane2005computational}. Due to the expensive computation of experiments or simulations, the data-driven surrogate model is frequently built \cite{kennedy2000predicting}. The surrogate model can be queried cheaply and is utilized to replace the original simulator. 
Popular specific implementations for surrogate models are Gaussian processes, radial basis function, support vector machines, neural networks, and Bayesian networks. 

Simulations of PDEs generally produce high-dimensional spatial or spatial-temporal field results. A spatial field is a function of many spatial variables. For example, the vector of velocity, temperature, pressure, or magnetics at points with 2D or 3D coordinates is a spatial field. 
Such situations require us to build a surrogate model between high-dimensional outputs and low-dimensional inputs, from a relatively small number of training samples. However, traditional statistical methods become inefficient when dealing with high-dimensional outputs, while learning an end-to-end model directly poses a huge challenge in terms of model complexity and computational cost \cite{higdon2008computer, xing2015reduced, xing2016manifold}. Therefore, it is of great significance to build a proxy of low-to-high mapping, based on a limited number of simulation samples. 

In this study, we propose the Extended Linear Model of Coregionalization (E-LMC), a novel framework that combines the neural network and the linear model of coregionalization (LMC) to act as a hybrid model for computationally expensive spatial field prediction defined in high-dimensional spaces. 
Neural network helps to disentangle the nonlinear features, whereas LMC linearly integrates base functions with latent processes to model the high-dimensional outputs. Our method can effectively capture the complicated correlations within spatial fields and provide accurate predictions. 
In theory, our method is a generalization of LMC for nonlinear problems.

Our work makes three main contributions to the high-dimensional spatial field prediction:
\begin{itemize}
\item We extend the LMC by introducing a general framework to linearize the highly complex and nonlinear outputs, allowing the LMC to generalize to nonlinear situations while preserving its original traceability and scalability.
\item We introduce the neural network to extract nonlinear correlations among spatial output variables, and present a simple training algorithm to alleviate overfitting.
\item We evaluate the proposed method on canonical physics-based simulations. Experimental results show a maximum of 40\% accuracy improvement over the original LMC and outperform the state-of-the-art spatial field models.
\end{itemize}

The remainder of this paper is structured as follows: Section \ref{Related Work} reviews the related work of the surrogate models in high-dimensional spatial field prediction; Section \ref{Method} presents the construction of our proposed E-LMC model; Section \ref{Experiments} describes the implementation details and illustrates our results; finally, conclusions for the whole study and future work are drawn in Section \ref{Conclusion}.

\section{Related Work}\label{Related Work}
\subsection{Learning the Spatial Field}
For physical simulations based on PDEs, the output is a spatial field, normally represented by its values on a regular grid. 
For example, as shown in Fig.~\ref{spatialField}, calculations of a magnetic field are usually performed using the finite element method (FEM) or finite difference method.
In this situation, the dimensionality of the output space is equal to the number of points in the spatial grid or the mesh, which means that these simulation methods are computationally intensive.
To conduct efficient analysis for engineering and scientific purposes, a practical way to reduce the high demand of computation is to incorporate surrogate models, which are essentially regression models, such as conservation kernels \cite{macedo2010learning} and hybrid physics-based data-driven surrogates \cite{shah2017reduced}. 

From the data-driven surrogate perspective, deep neural network (DNN) can be used to construct supervised models for numerical simulators. Multilayer perceptron (MLP), a typical representation of DNN, is capable of effectively finding a low-dimensional nonlinear manifold in high-dimensional data \cite{tripathy2018deep,nentwich2016application,takeishi2017learning}. However, these end-to-end neural networks or deep learning hybrid models require massive tuning and are prone to overfitting \cite{wilson2016deep, wilson2011gaussian}. Furthermore, deep learning models generally require a large amount of data, which is costly to obtain. Therefore, a stand-alone neural network architecture is not enough for modeling of the spatial fields.

\begin{figure}[ht]
\centering
\includegraphics[scale=0.23]{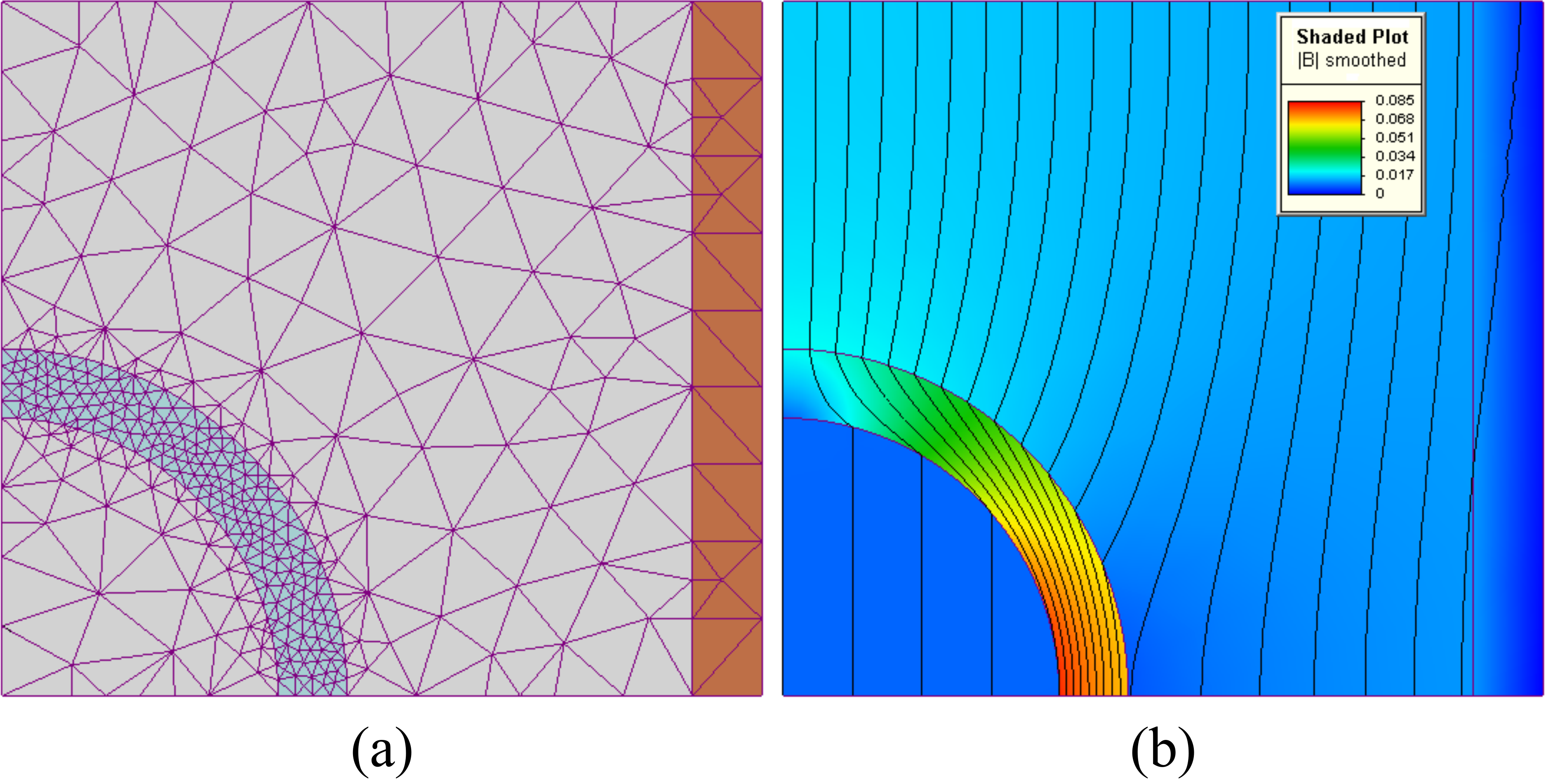}
\caption{An example of a 2D spatial field. (a) FEM mesh created prior to finding the magnetic field distribution of a certain geometry. Different colors indicate different material properties. (b) Output solution to the magnetic problem. The color shows the amplitude of the magnetic flux density.}
\label{spatialField}
\end{figure}

\subsection{Gaussian Process for Spatial Field Prediction}
As a nonparametric method that can adopt prior knowledge and provide uncertainty quantification, Gaussian process (GP) is frequently used as the data-driven surrogate model in kernel learning algorithms \cite{crevillen2018surrogate, zhe2019scalable, xing2021deep, hosseini2019large}. GP can alleviate overfitting and automatically capture the complexity of the functions underlying the data, from a limited number of training samples. In recent years, GP has been used for various kinds of applications, including optimization \cite{snoek2012practical}, face detection \cite{lu2015surpassing}, and collaborative filtering \cite{kim2016collaborative}.

Unfortunately, GP cannot be naturally extended to learn a function with multiple outputs, especially in the high-dimensional spatial fields. It is hard to apply GP to computational physics directly. 
Additionally, there are strong and complex correlations among the spatial outputs. If we ignore output correlations and simply treat each output independently (given inputs), it is likely to result in severe overfitting, particularly for small datasets. 
Therefore, multi-output GP models have been proposed to capture the output correlations. For example, the convolved GP models the covariance among the outputs through convolution operations \cite{higdon2002space, boyle2005dependent}. PCA-GP \cite{higdon2008computer}, Isomap-GP \cite{xing2015reduced}, and kPCA-GP \cite{xing2016manifold} assume a low-rank structure within the outputs and then model the mapping of this low-rank structure to the inputs through a linear combination of a group of fixed bases. 

Regardless of their effectiveness, for a general GP, the computational complexity is $\mathcal{O}(N^{3}d^{3})$, where $N$ is the number of training samples and $d$ is the output dimensionality \cite{mcfarland2008calibration}. As the number of required training samples grows drastically for a highly nonlinear response surface, the computation of GP-based surrogate modeling becomes less practical.
Therefore, the aforementioned surrogate models are either impractical for large spatial fields or inadequate to capture complicated correlations among very high-dimensional spatial outputs.

\subsection{Dimensionality Reduction in Spatial Field Prediction}
As discussed before, unlike ordinary datasets, data from PDEs are high-dimensional spatial fields with limited amount. An efficient approach is to perform dimensionality reduction on the output space.
Dimensionality reduction is defined as the mapping of the high-dimensional data into a low-dimensional representation \cite{jolliffe2016principal}. Lower dimension means less computation, less memory, and faster learning \cite{kong2015wind}. Dimensionality reduction can help extract the hidden information and analyze the underlying structure. It is also expected to improve the model accuracy by relieving overfitting \cite{tang2010dimensionality}.
Implementations of dimensionality reduction can be classified into linear and nonlinear methods.
Principal component analysis (PCA) is one of the most widely used and scalable linear dimensionality reduction techniques. The purpose of PCA is to decompose the original data into a few linearly uncorrelated variables based on an affine transformation. Lower-dimensional data are obtained while preserving as much of the data’s variation as possible by projecting each data point onto the first few principal components.

Higdon \emph{et al}. \cite{higdon2008computer} first used PCA combined with separate GP for spatial field prediction. Bayarri \emph{et al}. \cite{bayarri2007computer} proposed a similar method based on a wavelet decomposition. 
Unfortunately, PCA fails when the output space is not close to the linear subspace of the original space, while most physical or chemical processes in the real world generate nonlinear data.
Motivated by the work of Higdon, W. Xing \emph{et al}. \cite{xing2015reduced, xing2016manifold} employed nonlinear dimensionality reduction based on the Isomap \cite{tenenbaum1997mapping} and kernel PCA (kPCA) \cite{scholkopf1998nonlinear} to perform separate GP in a reduced-dimensional output space.

In recent decades, the potential of a neural network resides in nonlinearity, allowing the model to learn more powerful representations than the aforementioned traditional linear or nonlinear methods \cite{hinton2006reducing}.
It is known that a neural network with a linear activation function can learn the principal component representation of the input data \cite{bourlard1988auto}. 
There are also approaches to turn an autoencoder into a PCA, by generating an encoder layer using the PCA parameters and adding a decoding layer \cite{seuret2017pca, krahenbuhl2015data}. However, the weights of the autoencoder are not equal to the principal components, because they are generally not orthogonal.
Furthermore, in cases of low data volume, PCA offers better stability and generalization compared with single neural network, which requires big data and is prone to overfitting.
Therefore, we introduce PCA cooperating with neural network to reduce dimensionality of the spatial field and perform GP in the reduced-dimensional output space.

\section{Method}\label{Method}
\subsection{Statement of the Problem}
The predictions of the spatial field (usually produced from the simulation) can be considered as approximating a injective mapping $\pmb{\eta}:{\cal X}\rightarrow {\cal M}$, where ${\cal M}\subset \mathbb{R}^d$ is the permissible output space (response surface) and ${\cal X} \subset \mathbb{R}^l$ is the permissible input space. That is, $\pmb{\eta}(\pmb{\xi})= \mbox{\textbf{\textit{y}}}=(u(\textbf{\textit{x}}^{(1)};\pmb{\theta}),\dots,u(\textbf{\textit{x}}^{(d)};\pmb{\theta}))^T$, where $\pmb{\theta}$ is a vector of parameters and $\textbf{\textit{x}}$ is the spatial variable.
The objective of statistical emulation is to approximate the mapping $\pmb{\eta}$ for a given set of {\/\it training points\/}  $\mbox{\textbf{\textit{y}}}^{(j)}=\pmb{\eta}(\pmb{\xi}^{(j)})\in {\cal M}$, $j=1,\dots, m$. The corresponding inputs $\pmb{\xi}^{(j)} \in {\cal X}$ are referred to as {\/\it design inputs\/}. 

\subsection{Overview}
As shown in Fig.~\ref{E-LMC}, E-LMC consists of the MLP module and the LMC module. The first block of the MLP module disentangles the nonlinearity of the high-dimensional spatial outputs. Therefore, we capture the relationship among outputs and find a subspace of spatial output space that leads to an optimal representation of the LMC model response surface. Following that, we employ PCA and GP to implement the LMC module, which maps the inputs and the low-dimensional representations. Finally, with inverse PCA and the second block of the MLP module, we can reconstruct the linear and nonlinear characteristics to obtain the prediction.

We begin by introducing the basic GP model for scalar-valued spatial outputs. We then present the LMC module and an efficient implementation of it. Finally, we provide the MLP modules and propose the workflow of the E-LMC.

\begin{figure*}[ht]
\centering
\includegraphics[scale=0.45]{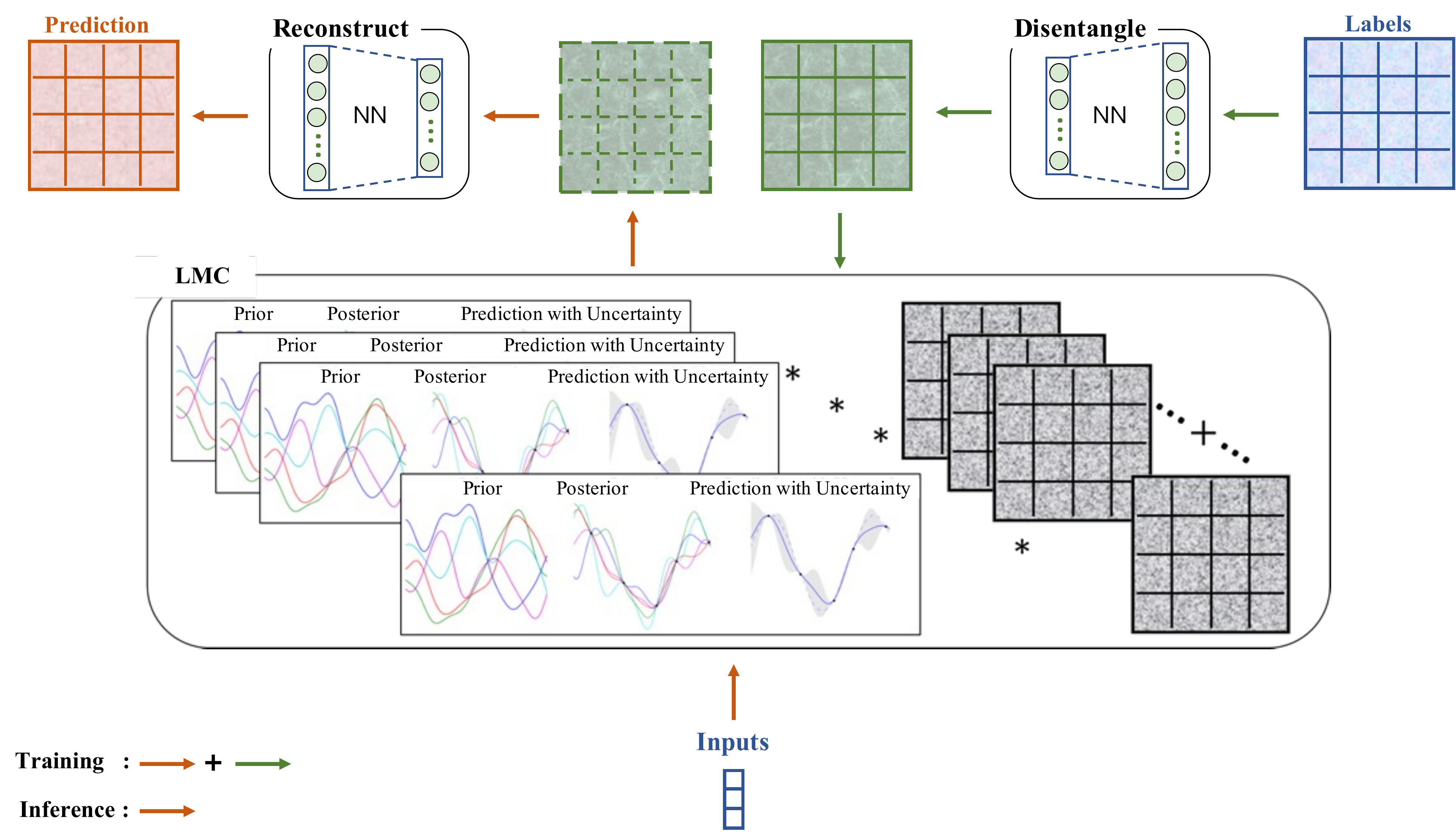}
\caption{Model architecture of the E-LMC.}
\label{E-LMC}
\end{figure*}

\subsection{Gaussian Process Regression}
Traditional surrogate models address this problem by using data generated from a few accurate simulations. For simplicity, consider the number of variables to be univariate. The {\/\it prior distribution\/} over the function $\eta(\pmb{\xi})$ in a GP model:

\begin{equation}
\label{scalaremb}
\eta(\pmb{\xi})| \pmb{\theta}\sim {\cal GP}\left(m(\pmb{\xi}),c(\pmb{\xi}, \pmb{\xi}'|\pmb{\theta})\right)
\end{equation}
where $\eta(\pmb{\xi})$ is distributed according to a GP, with mean and covariance functions $m_0(\pmb{\xi})$ and $c(\pmb{\xi}, \pmb{\xi}'|\pmb{\theta})$; $\pmb{\theta}$ is a vector of (usually unknown) parameters associated with the covariance function of a given functional form. 
The prior (\ref{scalaremb}) is updated using available data to obtain a posterior predictive GP distribution, with revised mean and covariance functions conditioned on  $\pmb{\theta}$ \cite{rasmussen2006gaussian}:
\begin{equation}
\label{postprednA}
\begin{array}{c}
\eta(\pmb{\xi})|{\bf t},\pmb{\theta}\sim {\cal GP}\left(m'(\pmb{\xi}|\pmb{\theta}),c'(\pmb{\xi},\pmb{\xi}'|\pmb{\theta})\right)
\vspace{2mm}\\
m'(\pmb{\xi}|\pmb{\theta})={\bf c}(\pmb{\xi})^T\textbf{C}(\pmb{\theta}) ^{-1}{\bf t}
\vspace{2mm}\\
c'(\pmb{\xi},\pmb{\xi}'|\pmb{\theta})=c(\pmb{\xi}, \pmb{\xi}'|\pmb{\theta})-{\bf c}(\pmb{\xi})^T\textbf{C}(\pmb{\theta}) ^{-1}{\bf c}(\pmb{\xi}')
\end{array}
\end{equation}
where ${\bf t}=(y_1,\hdots,y_M)^T$, $\textbf{C}(\pmb{\theta})=[C_{ij}]$, in which $C_{ij}=c(\pmb{\xi}_i, \pmb{\xi}_j|\pmb{\theta})$, $i,j=1,\hdots,M$, is the covariance matrix and ${\bf c}(\pmb{\xi})=(c(\pmb{\xi}_1, \pmb{\xi}|\pmb{\theta}),\hdots, c(\pmb{\xi}_M, \pmb{\xi}|\pmb{\theta}))^T$. The expected value $\mathbb{E}[\eta(\pmb{\xi})]$  is given by $m'(\pmb{\xi}|\pmb{\theta})$ and $c'(\pmb{\xi},\pmb{\xi}|\pmb{\theta})$ is the predictive variance. The parameters $\pmb{\theta}$ can be obtained from the maximum log-likelihood estimate \cite{xing2021new}.

\subsection{Linear Model of Coregionalization}
The basic GP can not be naturally extended to spatial outputs problems.
There are several ways to extend this approach to multivariate cases, which fall into the general framework of the linear model of coregionalization (LMC). 

Consider one of the outputs $\pmb{\eta} (\pmb{\xi})\in\mathbb{R}^{D}$, LMC assumes that the coordinates $\eta_d (\pmb{\xi})$, $d=1,\hdots,D$, of $\pmb{\eta} (\pmb{\xi})$ are linear combinations of $Q$ GPs (more generally, random functions):
\begin{equation}\label{LMC2b}
\eta_d (\pmb{\xi})=\sum _{q=1}^{Q}\sum _{i=1}^{R_q}a_{d,q}^iu_{q}^i(\pmb{\xi})
\end{equation}
where $a_{d,q}^i$ are scalar coefficients and $u_{q}^i(\pmb{\xi})$ are zero mean, unit-variance GPs that are related by $\operatorname {cov} (u_{q}^i(\pmb{\xi}),u^{i'}_{q'}(\pmb{\xi}'))=c_q(\pmb{\xi},\pmb{\xi}'|\pmb{\theta}_q)$ if $q=q'$ and $i=i'$, and $\operatorname {cov} (u_{q}^i(\pmb{\xi}),u^{i'}_{q'}(\pmb{\xi}'))=0$ otherwise. In above expressions, $\operatorname {cov} (\cdot,\cdot)$ denotes the covariance between the two arguments. $\pmb{\theta}_q$ is a vector of hyperparameters associated with each covariance function. For each fixed $q=1,\dots,Q$, the $u_{q}^i(\pmb{\xi})$, $i=1,\hdots,R_q$,  are independent (across $i$) and share the same correlation function $c_q(\pmb{\xi},\pmb{\xi}'|\pmb{\theta}_q)$, while  the $Q$ groups of functions $\{u_{q}^i\}_q, i=1,\hdots,R_q$, are independent of each other (i.e., across $q$). 
Setting ${\bf u}_q=(u_{q}^1(\pmb{\xi}), \hdots, u_{q}^{R_q}(\pmb{\xi}))^T$
and ${\bf A}_q=[a_{d,q}^{i}]\in\mathbb{R}^{D\times R_q}$, the matrix form of (\ref{LMC2b}) is:
\begin{equation}\label{LMCfullb}
\pmb{\eta} (\pmb{\xi})=\sum _{q=1}^{Q}\pmb{\eta}_q (\pmb{\xi})=\sum _{q=1}^{Q}{\bf A}_q {\bf u}_q(\pmb{\xi}).
\end{equation}
The cross-covariance matrix is $\operatorname {cov} \left(\pmb{\eta} (\pmb{\xi}),\pmb{\eta} (\pmb{\xi}')\right)=\sum _{q=1}^{Q}{\bf B}_{q} c_q(\pmb{\xi},\pmb{\xi}'|\pmb{\theta}_q)$, where ${\bf B}_{q}={\bf A}_q {\bf A}_{q}^T\in\mathbb{R}^{D\times D}$ are called {\/\it coregionalization matrices\/}, which encode the constant correlation among outputs for the $q$-latent process.
The complete set of hyperparameters in this model is $\{\pmb{\theta}_q,{\bf B}_q\}_q$.

The challenge for LMC is in estimating the coregionalization matrices characerized by ${\bf A}_{q}$, which leads to several special cases. 
To retain the full potential of LMC while reducing the computational cost and alleviating overfitting, Higdon \emph{et al}. \cite{higdon2008computer} performed PCA on the data ${\bf Y}=[{\bf y}_1,\hdots,{\bf y}_M]$ to approximate ${\bf A}_q$ using the first few eigenvectors.
PCA leads to subspace of $\mathbb{R}^D$ spanned by  eigenvectors ${\bf v}_q$, $q=1,\hdots,r$, of the sample covariance matrix ${\bf Y}{\bf Y}^T$ corresponding to the first $r<D$ eigenvalues (provided the eigenvalues are arranged in a non-increasing order). This suggests a  model of the form (\ref{LMCfullb}) with $Q=r$, $R_q=1$ and ${\bf A}_q ={\bf v}_q$.
The ${\bf u}_q$ are scalar, independent GPs $u_q(\pmb{\xi})$ that can be learned using the principal components obtained from the data \cite{xing2021new}.

\subsection{Multilayer Perceptron}
To harness stronger spatial output correlations and bring more scalability to LMC, we employ a neural network module to capture the nonlinear features. Specifically, we use MLP, which is composed of multiple fully-connected layers. Similar to an autoencoder, the output attempts to be as close as possible to its input and calculates an error. The error is used to adjust the weights by back-propagation and gradient descent algorithm.

The learning process for this module is composed of two processes: {\/\it disentangle\/} and {\/\it reconstruct\/}. The disentangle process occurs between the input layer and the LMC, extracting nonlinear features while leaving linear features. This is given by \eqref{encoder}.
\begin{equation}
\mathbf{h}=ReLU(\mathbf{b}+\mathbf{W}\mathbf{y}) \label{encoder}
\end{equation}
where $\mathbf{W}$ and $\mathbf{b}$ stand for the weight matrix and bias in the disentangle process; $\mathbf{y}$ denotes the input; $\mathbf{h}$ is the output which is the input of LMC; $ReLU$ (Rectified Linear Unit) is the activation function.

The reconstruct process entails the reconstruction of nonlinear features from the LMC. The process to obtain the output is given by \eqref{decoder}.
\begin{equation}
\mathbf{\hat{y}}=ReLU(\mathbf{b'}+\mathbf{W'}\mathbf{h'}) \label{decoder}
\end{equation}
where $\mathbf{W'}$ and $\mathbf{b'}$ stand for the weight matrix and bias in the reconstruct process; $\mathbf{h'}$ denotes the input which is the output of LMC; $\mathbf{\hat{y}}$ is the output; $ReLU$ (Rectified Linear Unit) is the activation function.

As previously mentioned, the reconstruct step tries to rebuild the disentangled data by comparing the obtained output with the original input. In order to evaluate this, we implement mean square error (MSE) as the loss function.

\subsection{Training and Inference of E-LMC}
So far, we have introduced the skeleton of our E-LMC model. The challenge is the optimization of model parameters.

The training and inference algorithms are shown in Algorithm~\ref{alg:Training} and Algorithm~\ref{alg:Testing}, respectively. Training includes the whole MLP module and the LMC module.
The computational complexity for our proposed method is $\Ocal(m^3+n^2)$ ($\Ocal(m^3)$ for GP and $\Ocal(n^2)$ for MLP, where $m$ is the number of training samples and $n$ is the the number of hidden units.)
Inference includes the reconstruct process of the MLP module and the LMC module.
$\mathbf{X}$ represents the low-dimensional inputs. $\mathbf{Y}$ represents the high-dimensional spatial outputs. $\mathbf{Y'}$ represents the high-dimensional spatial predictions from our model. $\mathbf{Z}$ represents the latent features/bases of the outputs. 
$W_D$ and $W_R$ are learnable weights in MLP. $\mu$ and $\Sigma$ are learnable means and covariances in GP.

\begin{algorithm}
  \caption{Training on E-LMC}
  \label{alg:Training}
  initialization on MLP\;
  \For{number of training iterations}{
    \For{number of batches of the training dataset}{
      $\mathbf{MLP}(\mathbf{Y}, W_D) \rightarrow \mathbf{M}$ \\
      $\mathbf{PCA}(\mathbf{M}) \rightarrow \mathbf{Z}$ \\
      $\mathbf{InversePCA}(\mathbf{Z}) \rightarrow \mathbf{M'}$ \\
      $\mathbf{MLP}(\mathbf{M'}, W_R) \rightarrow \mathbf{Y'}$ \\
      Update MLP parameters $W_D$ and $W_R$ \\
      through gradient descent.
      }
    }

  initialization on GP\;
  \For{number of training iterations}{
    $\mathbf{GP}(\mathbf{X}, \mu, \Sigma) \rightarrow \mathbf{Z}$ \\
    Update the mean and covariance of GP \\
    through gradient descent.
    }
\end{algorithm}

\begin{algorithm}
  \caption{Inference on E-LMC}
  \label{alg:Testing}
  \For{number of testing iterations}{
    \For{number of batches of the testing dataset}{
      $\mathbf{GP}(\mathbf{X}, \mu, \Sigma) \rightarrow \mathbf{Z}$ \\
      $\mathbf{InversePCA}(\mathbf{Z}) \rightarrow \mathbf{M'}$ \\
      $\mathbf{MLP}(\mathbf{M'}, W_R) \rightarrow \mathbf{Y'}$
      }
    }
\end{algorithm}

\section{Experiments}\label{Experiments}

\subsection{Datasets}
In order to assess the performance of E-LMC, two representative real-world physics-based datasets were used.

The first is the cantilever beam dataset. This dataset is about the structural topology optimization of a cantilever beam \cite{bendsoe2003topology}, shown in Fig.~\ref{canti}(a). Structural topology optimization is typically associated with a large number of simulation parameters and has a high computational cost. 
Cantilever beams are material structures that have a maximum stiffness when bearing forces from the right side.
There are three simulation parameters: the location of point load $P_{1}$, the angle of point load $P_{2}$, and the filter radius $P_{3}$ \cite{bruns2001topology}. 
The simulation triplet $P=(P_{1}\in[-20, 20], P_{2}\in[0, \pi], P_{3}\in[1.1, 2.5])$ is the input, while the $40 \times 80$ image presents the spatial field output \cite{xing2020shared}.

The second is the metal melting front dataset. A square cavity containing liquid and solid contents is subjected to a temperature difference between the left and right boundaries \cite{wolff1988solidification}. 
The temperature boundary conditions are shown in Fig.~\ref{canti}(b). The temperature ranges from $T_h$ to $T_c < T_h$ along the edges.
The fluid and solid phases are considered to be separate domains sharing a moving melting front. The liquid velocity is recorded on a $100 \times 100$ square spatial grid as outputs, and the corresponding inputs are three-dimensional.

We consider the cantilever beam and metal melting front as effective demonstrations for E-LMC because of their high-dimensional spatial outputs.

\begin{figure}[htbp]
\centering
\includegraphics[scale=0.28]{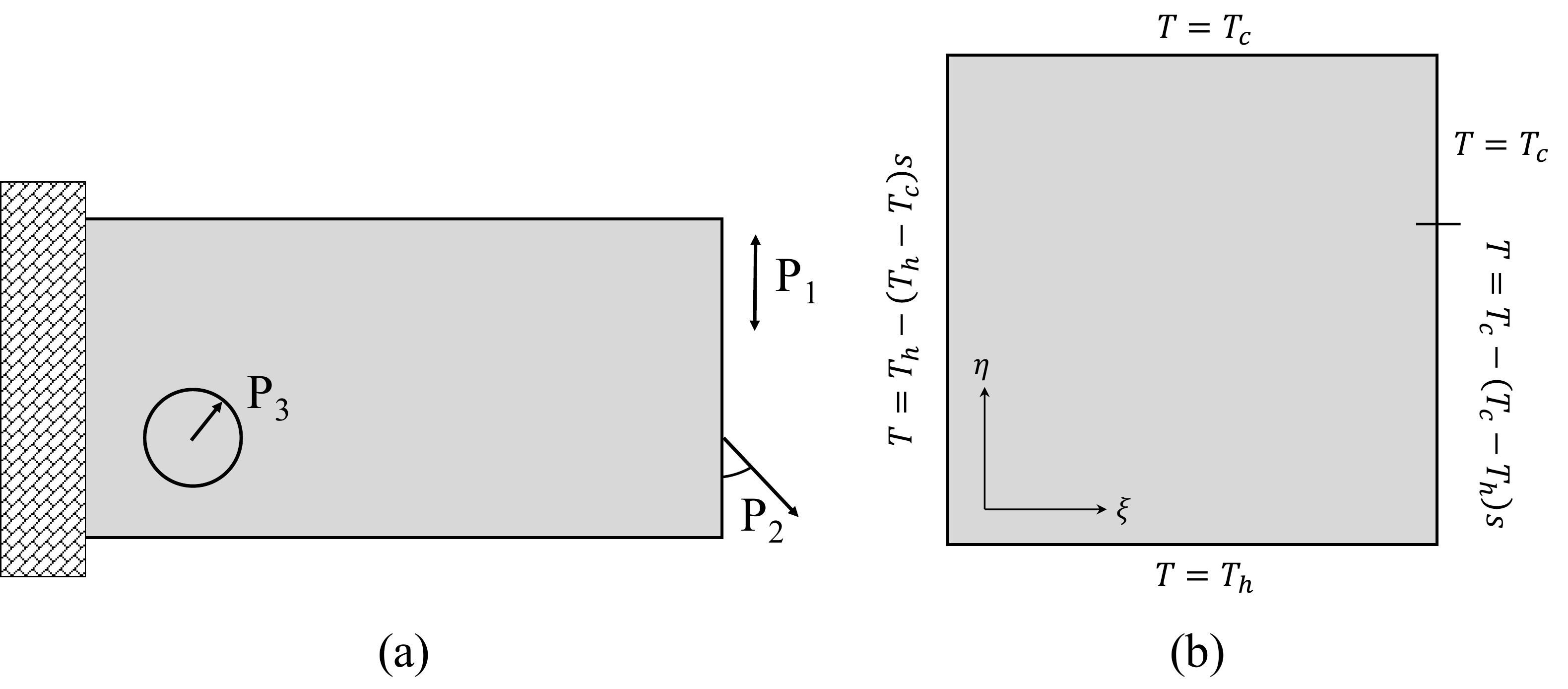}
\caption{(a) The geometry and external forces for the cantilever beam. (b) Temperature boundary conditions for the metal melting simulation.}
\label{canti}
\end{figure}

\subsection{Baselines}
In order to comprehensively evaluate the performance of E-LMC, we compare with the vanilla MLP model and the GP-based models:
\begin{itemize}
\item \textbf{MLP}: This end-to-end deep learning model using fully-connected neural networks predicts the high-dimensional spatial outputs directly from the low-dimensional inputs.
\item \textbf{LMC (PCA-GP)} \cite{higdon2008computer}: This multi-output GP regression approach generates the outputs via a linear combination of fixed bases. PCA is used to derive the bases from the training samples.
\item \textbf{Isomap-GP} \cite{xing2015reduced}: This method exploits patterns in the permissible output space using manifold learning. Isomap and kernel Isomap are used to significantly reduce the dimensionality of the output space to construct a GP emulator efficiently.
\item \textbf{kPCA-GP} \cite{xing2016manifold}: This is a cogenetic method similar to Isomap-GP. The kernel PCA and diffusion maps are developed to reduce the dimensionality of outputs.
\item \textbf{HOGP} \cite{zhe2019scalable}: This high-order GP regression model can flexibly capture complex correlations among the outputs and naturally scale up to high-dimensional outputs.
\end{itemize}

\subsection{Experimental Settings}
For E-LMC, we organize the outputs from the cantilever beam and metal melting front datasets into images with sizes of $40 \times 80$ and $100 \times 100$, respectively. The model is built based on the PyTorch and GPyTorch frameworks. We adapt ten fully-connected layers for the MLP module, and the number of units for each hidden layer is $\left\{4096, 8192, 4096, 3200, 3200, 4096, 8192, 4096\right\}$. 
We use the RBF squared exponential kernel for GP in the LMC module. We employ the latent features/bases $rank$, and vary $rank$ from $\left\{1, 2, 5, 10\right\}$. 
The loss function for GP is the marginal log likelihood.
We use Adam as the optimizer for MLP and GP, with learning rate as 0.001 and set the batch size as 64.

For MLP, if we use the same large number of parameters as the MLP module in E-LMC, this vanilla MLP suffers from severe overfitting. Therefore, we manually adjust and set the number of units for each hidden layer to $\left\{8, 16, 32, 64, 128, 256, 512, 1024, 2048\right\}$.
The optimizer is Adam, with a learning rate of 0.001 and a batch size of 32.

For HOGP, PCA-GP, Isomap-GP, and kPCA-GP, we vary the number of latent features $rank$ from $\left\{1, 2, 5, 10\right\}$ as well. We use the RBF squared exponential kernel for the GP blocks and the same initialization for the kernel parameters and the inverse variance. These four baselines are implemented with MATLAB 2019. We use Adam as the optimizer with learning rate as 0.001. 

From the cantilever beam dataset, we randomly select $\left\{10, 50, 100, 500, 1000\right\}$ samples for training, and the remaining data for testing. From the metal melting front dataset, we randomly select $\left\{10, 50, 100\right\}$ samples for training, and the remaining data for testing.
We use the mean square error (MSE) as metrics in our experiments.
We report the average testing results over five random shuffles on the datasets.

\begin{figure}
\centering
\includegraphics[scale=0.084]{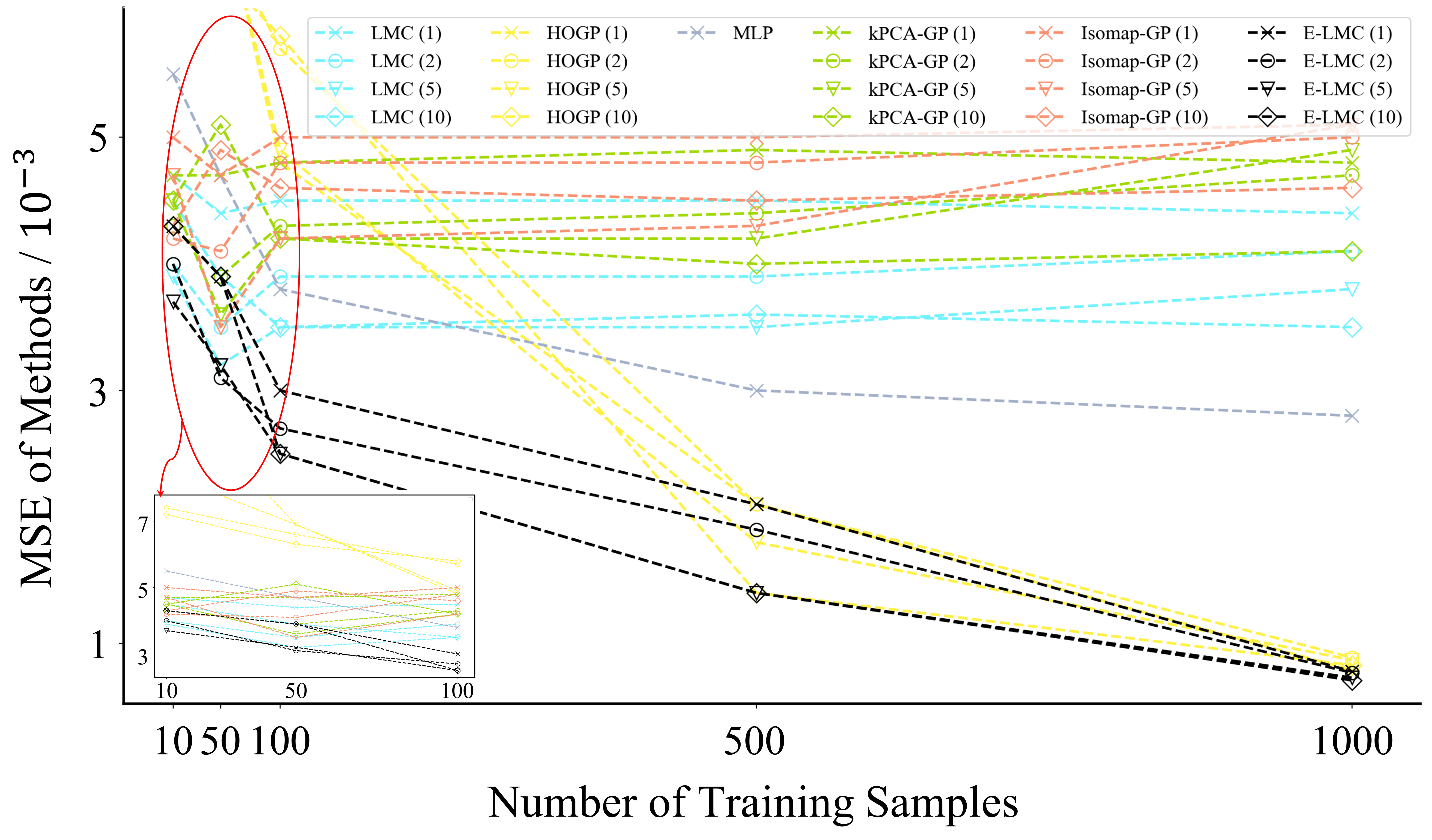}
\caption{The MSE on cantilever beam dataset w.r.t different sizes of training set. Numbers in the legend indicate the number of latent features for E-LMC, LMC (PCA-GP), HOGP, kPCA-GP, and Isomap-GP.}
\label{MSE_RANK_ntr_canti_total}
\end{figure}

\begin{figure}
\centering
\includegraphics[scale=0.24]{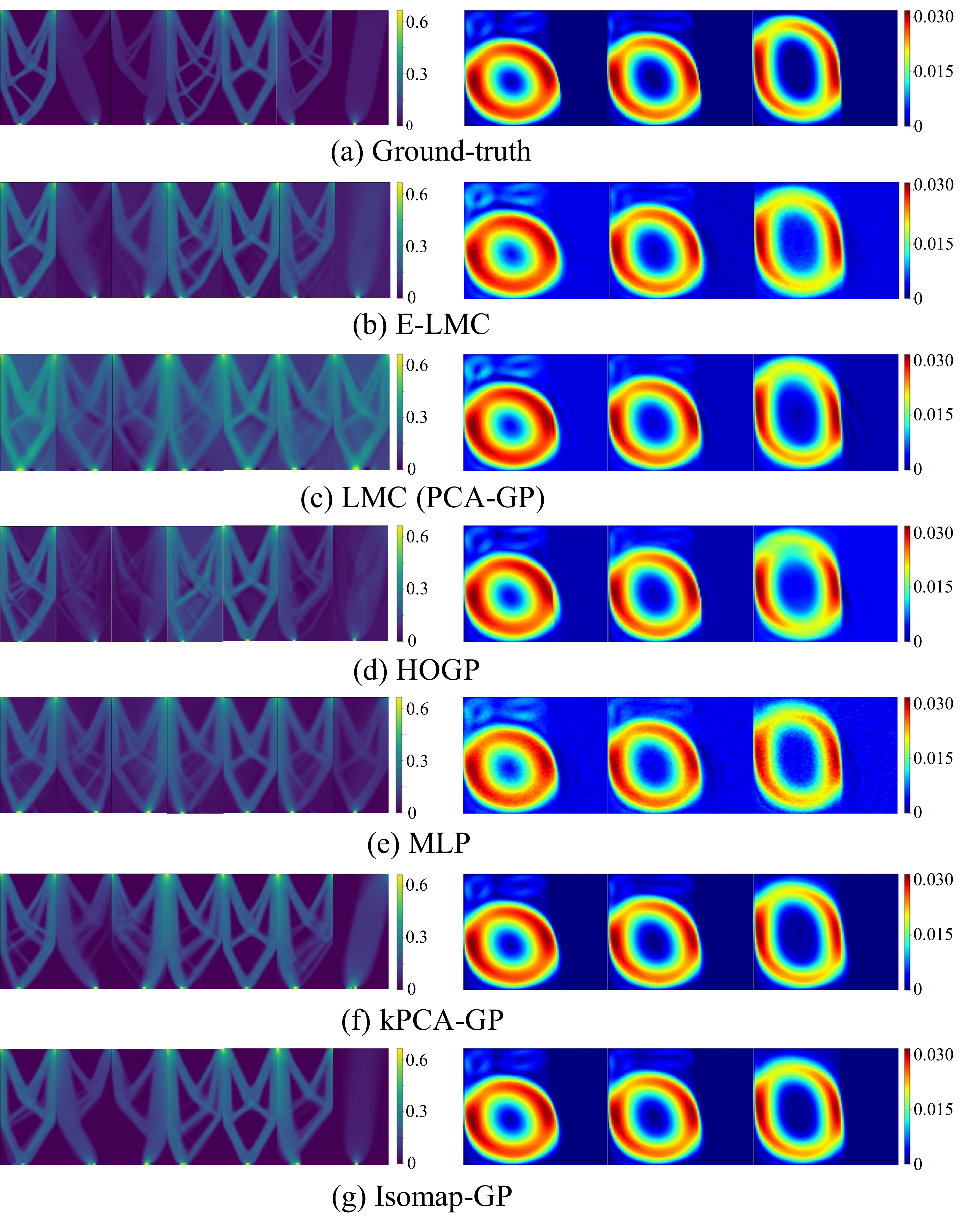}
\caption{The predicted topology structures for cantilever beam dataset (left) and the predicted metal conditions for metal melting front dataset (right). The number of latent features for E-LMC, LMC (PCA-GP), HOGP, kPCA-GP, and Isomap-GP were set to 10. Models were trained on 1000 training samples for cantilever beam dataset and 100 training samples for metal melting front dataset.}
\label{canti_metal_pred_image}
\end{figure}

\begin{table*}
\captionsetup{labelsep=newline, justification=centering}
\renewcommand\arraystretch{1.25}
\caption{Comparison Results with Baselines on Datasets ($rank = 10$)}
\begin{center}
\begin{tabular}{cccccccc} 
\toprule
\multirow{2}{*}{Dataset}                                                                           & \multirow{2}{*}{Training Samples} & \multicolumn{6}{c}{MSE of Methods}                                                                                        \\ 
\cline{3-8}
                                                                                                   &                                   & \textit{\textbf{E-LMC}} & \textit{LMC} & \textit{HOGP} & \textit{MLP} & \textit{kPCA-GP} & \textit{Isomap-GP}  \\ 
\hline
\multirow{5}{*}{\begin{tabular}[c]{@{}c@{}}Cantilever Beam\\(MSE base $10^{-3}$)\end{tabular}}     & $10$                              & \bm{$4.3$}                    & $4.5$                  & $7.2$         & $5.5$        & $4.5$            & $4.3$               \\
                                                                                                   & $50$                              & \bm{$3.9$}                    & $3.9$                  & $6.3$         & $4.7$        & $5.1$            & $4.9$               \\
                                                                                                   & $100$                             & \bm{$2.5$}                    & $3.5$                  & $5.8$         & $3.8$        & $4.2$            & $4.6$               \\
                                                                                                   & $500$                             & \bm{$1.4$}                    & $3.6$                  & $1.4$         & $3.0$        & $4.0$            & $4.5$               \\
                                                                                                   & $1000$                            & \bm{$0.71$}                   & $3.5$                  & $0.83$        & $2.8$        & $4.1$            & $4.6$               \\ 
\hline
\multirow{3}{*}{\begin{tabular}[c]{@{}c@{}}Metal Melting Front\\(MSE base $10^{-6}$)\end{tabular}} & $10$                              & \bm{$14$}                     & $22$                   & $64$          & $82$         & $20$             & $18$                \\
                                                                                                   & $50$                              & \bm{$4.1$}                    & $4.2$                  & $5.8$         & $8.0$        & $4.9$            & $4.5$               \\
                                                                                                   & $100$                             & \bm{$2.1$}                    & $3.9$                  & $4.9$         & $3.8$        & $3.5$            & $3.5$               \\
\bottomrule
\end{tabular}
\label{tab1}
\end{center}
\end{table*}

\begin{figure*}
\centering
\includegraphics[scale=0.18]{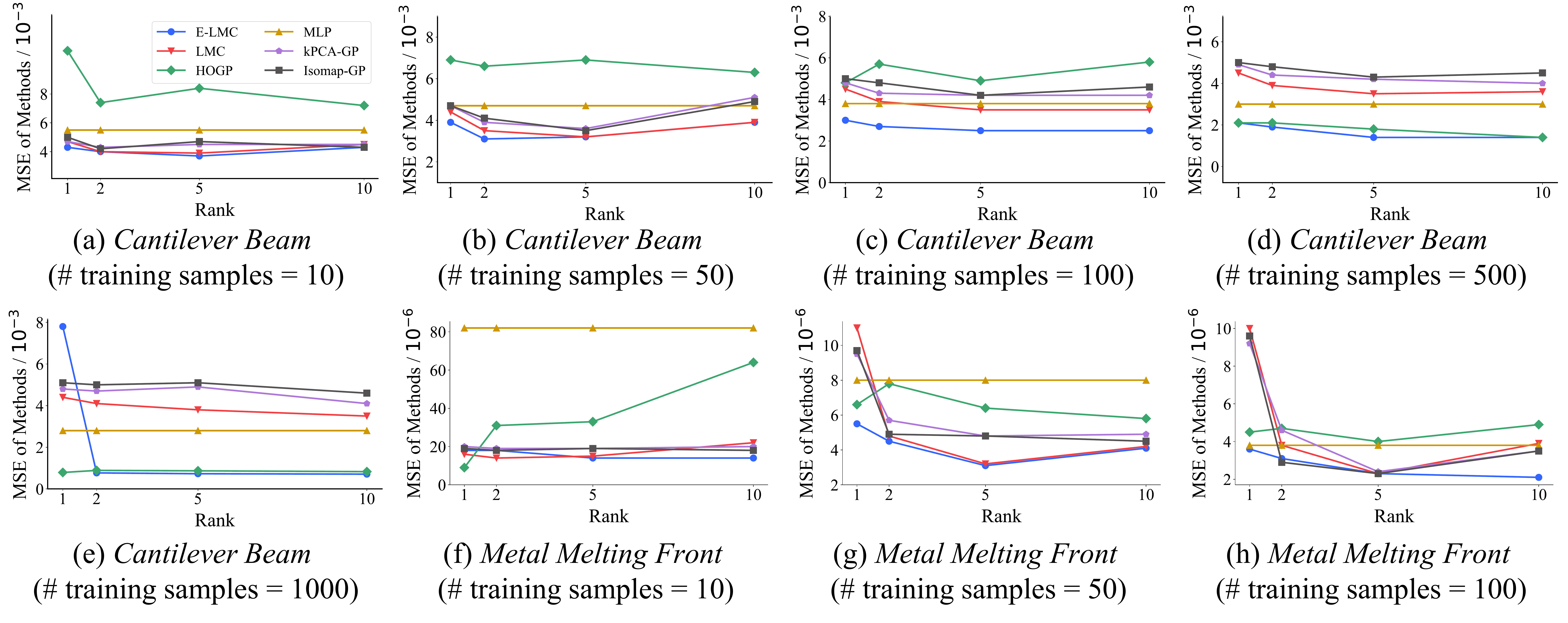}
\caption{The y-axis represents the MSE of all the methods on 2 small datasets. The x-axis represents the number of latent features for E-LMC, LMC (PCA-GP), HOGP, kPCA-GP, and Isomap-GP.}
\label{MSE_RANK}
\end{figure*}

\subsection{Results}
As shown in Fig.~\ref{MSE_RANK_ntr_canti_total}, E-LMC achieves significant improvements over all the baseline methods by a substantial margin in MSE.
Generally, the performances of almost all the methods are close when the size of the training set is small. When the training size increases over 100, E-LMC, HOGP, and MLP exhibit considerable benefits over all competing methods. The explanation for this phenomenon might be that the training set is too small to provide useful information at the start, leading to poor prediction performance. When the size of training samples increases, E-LMC can capture complicated output correlations and show superior prediction performance. 
LMC has the nonparametric stability of kernel methods and could relieve overfitting, while MLP is overparameterized and could automatically discover meaningful representations in high-dimensional data \cite{wilson2016deep}.
In summary, E-LMC has the stability in GP surrogate models for small datasets and the data-driven property in neural network models for big datasets.

The comparison results for 10 latent features on two datasets are presented in Table~\ref{tab1}. Specifically, E-LMC outperforms the state-of-the-art algorithm by 14.4\% on 1000 training samples from cantilever beam dataset and 40\% on 100 training samples from metal melting front dataset.

To allow for a fine-grained analysis, we visualize 7 predictions from the cantilever beam dataset and 3 predictions from the metal melting front dataset by all the methods (the number of latent features is set to 10, and the number of training samples is set to 1000), as well as the ground-truth. All of the chosen samples are representative of the corresponding datasets.
As illustrated in Fig.~\ref{canti_metal_pred_image}, on the cantilever beam dataset, the structures predicted by E-LMC are closest to the ground-truth. 
Specifically, predictions by LMC (PCA-GP) usually have blurred structures. 
HOGP, kPCA-GP, and Isomap-GP yield clearer local structures. MLP cannot deal with a variety of forms, and the local details sometimes diverge greatly from the ground-truth (see the second, third, and seventh structures in Fig.~\ref{canti_metal_pred_image}e). This vividly demonstrates that the neural network is prone to overfitting between high-dimensional outputs and low-dimensional inputs from a relatively small dataset. In constrast, E-LMC alleviates the overfitting issue by combing neural network with LMC.
On the metal melting front dataset, all models predict decent performances and are close to the ground-truth.

Fig.~\ref{MSE_RANK} shows the prediction MSE of all methods with respect to different latent features (i.e., $rank$) on different sizes of training sets. It is worth noticing that HOGP and MLP are not very stable when the number of latent functions is set to one (see Fig. 5a, 5b, 5c, 5f). For example, in Fig. 5f, the average MSE of MLP turns out to be far greater than all the other methods. In contrast, our E-LMC performs effectively even when only taking one latent feature.

\section{Conclusion}\label{Conclusion}
In this paper, we propose the extended linear model of coregionalization (E-LMC) for spatial field prediction. We introduce an invertible neural network to linearize the highly complex and nonlinear correlations among spatial output variables and cooperate with LMC to map low-dimensional inputs to high-dimensional outputs, based on a small dataset.
Real-world experiments demonstrate that E-LMC can exploit the spatial correlations effectively, showing a maximum improvement of about 40\% over other state-of-the-art spatial field prediction models.

In the future, the physical interpretation of the proposed models should be further investigated. Another improvement that could be made is to train neural networks and Gaussian process together for improved prediction accuracy, though we have got a decent performance by training them seperately in this study.

\section*{Acknowledgment}
The authors thank the National Natural Science Foundation of China (Grant No. 12004024).
The authors would like to acknowledge the computation supports from Shucheng Ye and Yinpeng Wu of Beihang University.

\bibliographystyle{IEEEtran}
\bibliography{reference}

\end{document}